\documentclass{article}
\usepackage{spconf,amsmath,epsfig}
\usepackage{amssymb}
\usepackage{booktabs}
\usepackage{arydshln}
\usepackage{multirow}
\usepackage{times}
\usepackage{enumerate}
\usepackage[pagebackref=false,breaklinks=true,colorlinks]{hyperref}
\let\OLDthebibliography\thebibliography
\renewcommand\thebibliography[1]{
	\OLDthebibliography{#1}
	\setlength{\parskip}{0pt}
	\setlength{\itemsep}{0pt plus 0.3ex}
}
\usepackage{url}
\begin{document}\sloppy
	
	% Example definitions.
	% --------------------
	\def\x{{\mathbf x}}
	\def\L{{\cal L}}

	% Title.
	% ------
	\title{Inharmonious Region Localization}
	%
	% Single address.
	% ---------------
	%\copyrightnotice{978-1-6654-3864-3/21/\$31.00 ©2021 IEEE} 
	\name{Jing Liang, Li Niu\textsuperscript{*}\thanks{\textsuperscript{*}Corresponding author.},  Liqing Zhang}
	\address {MoE Key Lab of Artificial Intelligence, Department of Computer Science and Engineering \\
		Shanghai Jiao Tong University, Shanghai, China \\ 
		\{leungjing, ustcnewly\}@sjtu.edu.cn, zhang-lq@cs.sjtu.edu.cn.}

	\maketitle
	\begin{abstract}
		The advance of image editing techniques allows users to create artistic works, but the manipulated regions may be incompatible with the background. Localizing the inharmonious region is an appealing yet challenging task. Realizing that this task requires effective aggregation of multi-scale contextual information and suppression of redundant information,  we design novel Bi-directional Feature Integration (BFI) block and Global-context Guided Decoder (GGD) block to fuse multi-scale features in the encoder and decoder respectively. We also employ Mask-guided Dual Attention (MDA) block between the encoder and decoder to suppress the redundant information.  
		Experiments on the image harmonization dataset demonstrate that our method achieves competitive performance for inharmonious region localization. The source code is available at \href{https://github.com/bcmi/DIRL}{https://github.com/bcmi/DIRL}.
	\end{abstract}

	
	\vspace{-0.2cm}
	\section{Introduction}
	\label{sec:intro}
	
\begin{figure}[t]
	
		\centering
		\includegraphics[scale=0.13]{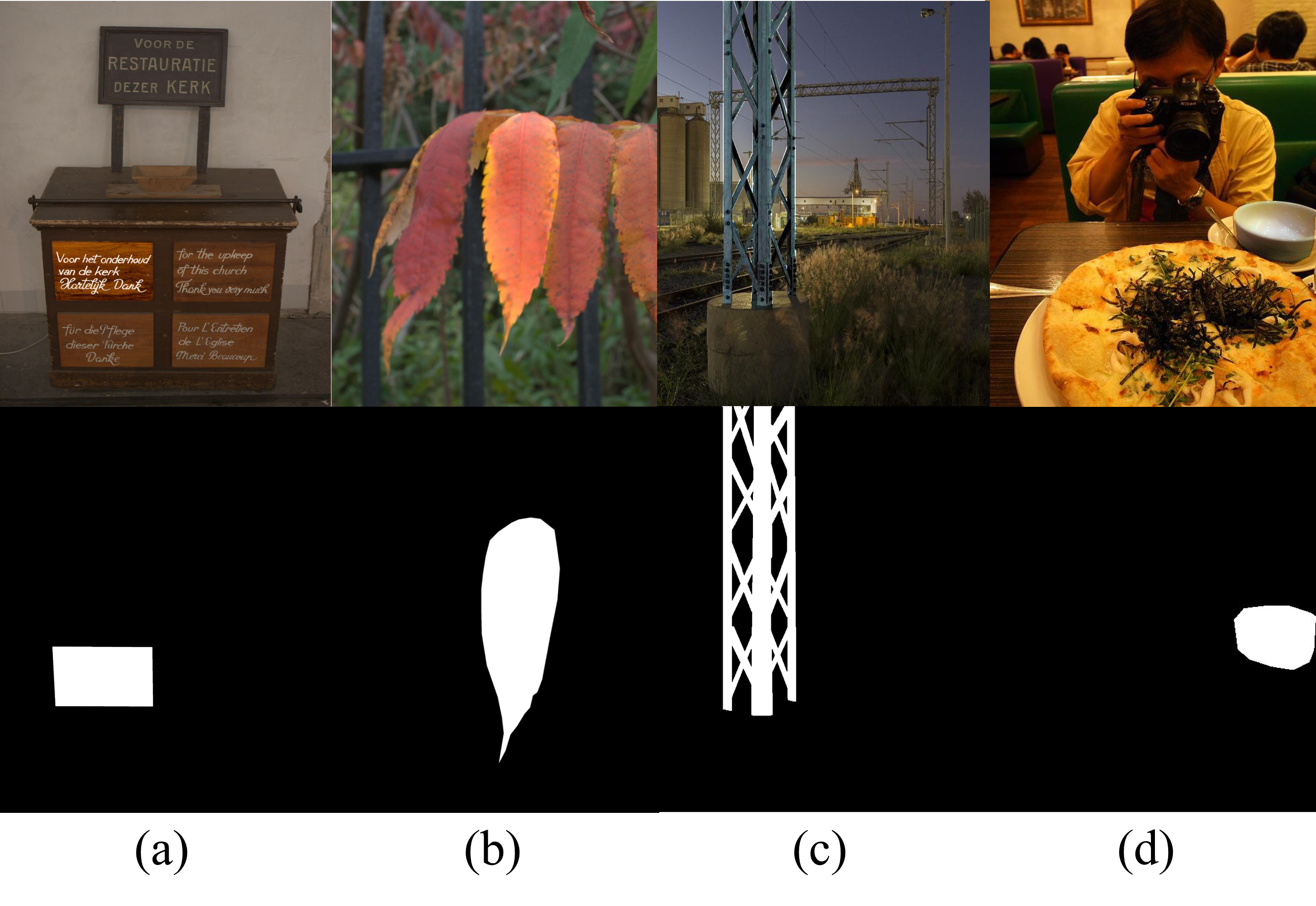}
		\caption{Examples of inharmonious images (top row) and their inharmonious region masks (bottom row).}
		\label{fig:inharmonious_clues}
	\end{figure}

	With the prevalence of image capturing devices and the advance of image editing techniques (\emph{e.g.}, appearance adjustment,  copy-paste), users can easily edit existing images to produce synthetic artistic works. However, in these synthetic images, some manipulated regions may look incompatible with the background due to inconsistent color and lighting statistics as shown in Fig.~\ref{fig:inharmonious_clues}, which significantly degrades the realism and quality of synthetic images. 
	
	Given a synthetic image, we refer to the region incompatible with the background as inharmonious region in terms of the color or lighting consistency. The task of inharmonious region localization aims to localize the inharmonious region in a synthetic image, after which users can manually adjust the inharmonious region or utilize automatic image harmonization techniques~\cite{tsai2017deep,cun2020improving,cong2019deep} to improve the quality of synthetic image. 
	
	One critical challenge of inharmonious region localization is that one region needs to be compared with multi-scale nearby regions to determine whether it is an inharmonious region, but integrating multi-scale contextual information will inevitably introduce redundant information~\cite{pang2020multi,zhang2018bi}. 
	Therefore, how to fuse multi-scale features and avoid redundant information is the key for successfully localizing inharmonious region.
	We adopt the popular encoder-decoder structure due to its great success in many computer vision tasks~\cite{gu2020hard,zhang2017amulet,isola2017image}. To facilitate thorough multi-scale feature fusion, we propose a novel Bi-directional Feature Integration (BFI) block for the encoder and a novel Global-context Guided Decoder (GGD) block for the decoder. To suppress redundant information, we insert Mask-guided Dual Attention (MDA) blocks between the encoder and the decoder. 
	
	Our whole network structure is illustrated in Fig.~\ref{fig:network}. It is well-known that shallow (\emph{resp.}, deep) encoder blocks produce high-resolution low-level (\emph{resp.}, low-resolution high-level) features. First, we design a novel Bi-directional Feature Integration (BFI) block to fuse multi-scale encoder features. 
	Inspired by \cite{pang2020multi}, we aggregate the adjacent encoder features because this can encourage sharing more relevant information and avoid the interference caused by large resolution differences~\cite{pang2020multi}. Each BFI block takes two or three adjacent encoder features as input to go through a top-down stream and a bottom-up stream in parallel for information propagation and aggregation. Then, the output of each BFI block is fed into a Mask-guided Dual Attention (MDA) block which consists of spatial attention and channel attention as depicted in CBAM~\cite{woo2018cbam}. To make the network focus on inharmonious region and alleviate the attention drift problem~\cite{cheng2017focusing}, we use inharmonious mask to supervise the learnt spatial attention map. Finally, the outputs from MDA blocks are fed into the decoder. Note that the encoder blocks and decoder blocks are connected in a UNet-like fashion. However, in our decoder block, we not only fuse the connected encoder feature and the previous decoder feature, but also inject the global contextual information (the output from the highest-level MDA block) into each decoder block with a shortcut, yielding our Global-context Guided Decoder (GGD) block. We name our method as Deep Inharmonious Region Localization (DIRL) network.

	Inharmonious region localization is related to previous image manipulation localization tasks. However, some methods in this field rely on strong prior knowledge such as camera color filter array inconsistency~\cite{chen2017image,hu2016effective} and compression artifacts inconsistency~\cite{ZHAO2013158,6151134}. Other methods~\cite{Wu_2019_CVPR,bappy2019hybrid} attempt to detect arbitrary manipulation type (\emph{e.g.}, copy-move, splicing, removal, enhancement, \emph{etc.}) and it is unclear which clues are used to identify the manipulated region. Different from the above methods, we focus on localizing the inharmonious region which is incompatible with the background due to color and lighting inconsistency.

	The main contributions of this paper can be summarized as follows: (1) This is the first work focusing on inharmonious region localization task; (2) We propose a new network with novel BFI block and GGD block for thorough multi-scale feature fusion and redundancy suppression; (3) Extensive experiments on iHarmony4 dataset demonstrate that our proposed method can outperform the state-of-the-art methods from related fields.
	
	\vspace{-0.2cm}
	\section{Related Works}\label{sec:relatedworks}
	\subsection{Image Manipulation Localization}

	Inharmonious region localization bears some resemblance to image manipulation localization.
	Image manipulation localization covers a wide range of manipulation types including splicing, copy-move, removal, enhancement, and so on.  
	
	Earlier image manipulation localization works heavily depend on the prior knowledge about the inconsistency between the manipulated region and the background, like camera color filter array inconsistency~\cite{chen2017image,hu2016effective},
	JPEG compression artifact inconsistency~\cite{ZHAO2013158,6151134}, or blur type inconsistency~\cite{bahrami2015blurred}.
	With the advance of deep learning, a bunch of general image manipulation localization approaches~\cite{Wu_2019_CVPR,bappy2019hybrid,bappy2017exploiting,Salloum_2018,kniaz2019point,9102825} have been proposed, but it is unclear which clues are used to localize the manipulated region. The potential clues include the inconsistency from many aspects such as perspective, geometry, noise pattern, frequency characteristics, color, lighting, \emph{etc}. In contrast, this is the first work focusing on inharmonious region localization, which aims to localize the inharmonious region based on the clues of color and lighting inconsistency. 
	
	\vspace{-0.2cm}
	\subsection{Image Harmonization}
	Image harmonization task, which aims to adjust the inharmonious region to make it compatible with the background, is also related to inharmonious region localization. 
	Tsai \emph{et al.}~\cite{tsai2017deep} proposed the first end-to-end framework to predict the harmonized images. Cong \emph{et al.}~\cite{cong2019deep} proposed to pull the domain of inharmonious region to that of background using a domain verification discriminator. In S2AM~\cite{cun2020improving}, an attention module was introduced to learn the foreground and background feature separately. Nevertheless, most of them require the inharmonious region mask as input, otherwise the performance will be significantly degraded.  S2AM~\cite{cun2020improving} considered the case without input mask and could predict the inharmonious mask. But mask prediction is not the focus of S2AM~\cite{cun2020improving} and the predicted mask is far from satisfactory. 
	In contrast, our work focuses on inharmonious region localization and could predict high-quality mask.  
	
	\vspace{-0.2cm}
	\subsection{Multi-Scale Feature Fusion}
	Recently, multi-scale feature fusion has achieved remarkable success in many computer vision fields like object detection~\cite{lin2017feature}, salient object detection~\cite{zhang2017amulet,pang2020multi}, instance segmentation~\cite{liu2018path}. However, previous methods mainly fused multi-scale features only in the encoder~\cite{lin2017feature, gu2020context} or only in the decoder~\cite{zhang2017amulet,ronneberger2015u}. Besides, most of them did not consider the induced redundant information when integrating multi-scale features. Instead, we fuse multi-scale features in both encoder and decoder while suppressing the redundancy. 
	
	The module closest to our Bi-directional Feature Integration (BFI) block is Aggregate Interaction Module (AIM) in \cite{pang2020multi}, which also attempts to fuse adjacent encoder features. However, AIM simply merges neighboring encoder features without holistic information exchange. Instead, our BFI accomplishes bi-directional information propagation and aggregation to facilitate holistic information exchange.

	\begin{figure*}[!ht]
		\centering
		\includegraphics[scale=0.28]{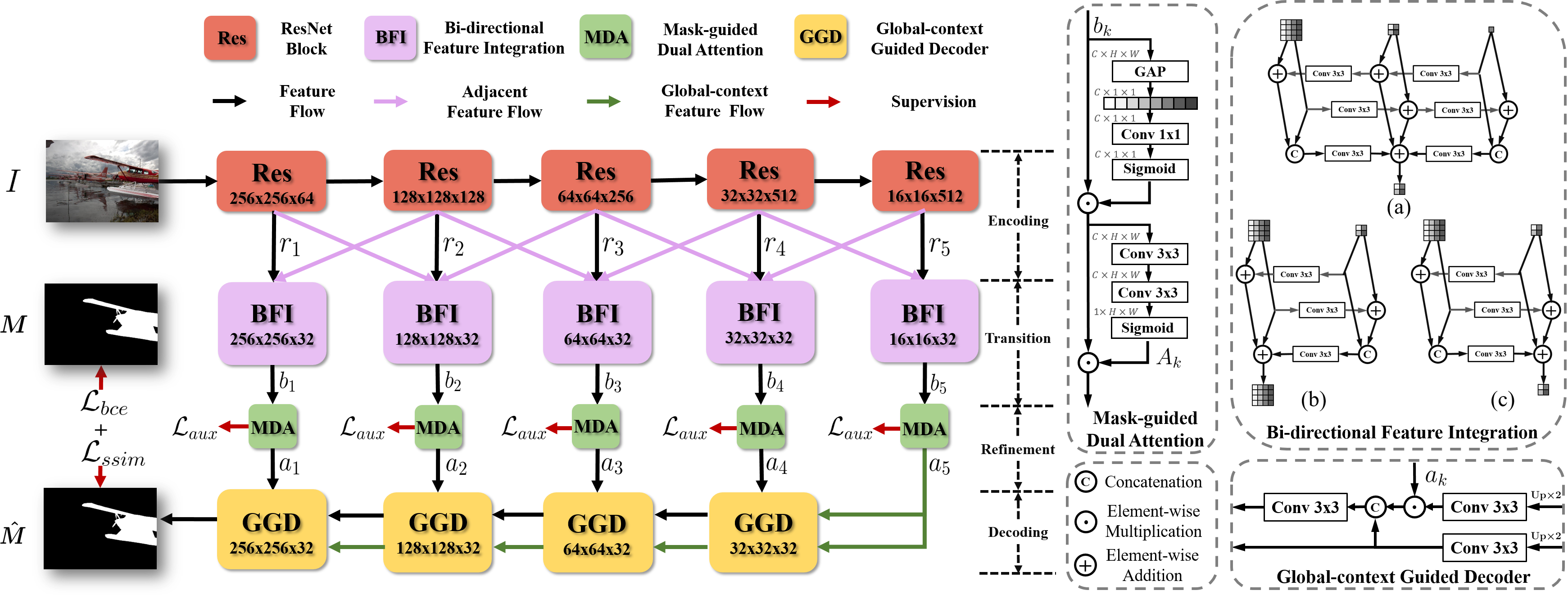}
		\caption{The left part shows our DIRL network which consists of four stages: encoding stage, transition stage, refinement stage, and decoding stage. The right part elaborates our proposed Bi-directional Feature Integration (BFI) block, Mask-guided Dual Attention (MDA) block, and Global-context Guided Decoder (GGD) block.  ``GAP" means global average pooling, ``Conv k x k" means the convolution layer with kernel size $k$, and ``Sigmoid" is the sigmoid activation function. }
		\label{fig:network}
	\end{figure*}

	\vspace{-0.2cm}
	\section{Our Method}\label{sec:method}

	\subsection{Overview}\label{sec:overview}
	Given a synthetic image $I \in \mathbb{R}^{H \times W \times 3}$, we aim to predict its inharmonious region mask $\hat{M} \in \mathbb{R}^{H \times W \times 1}$. 
	As illustrated in Fig.~\ref{fig:network}, the whole network consists of four stages: \textbf{encoding stage, transition stage, refinement stage, and  decoding stage}. In the encoding stage, we utilize five basic res-blocks to obtain multi-scale encoder features $\{r_k|_{k=1}^5\}$, where the first four blocks are adopted from ResNet34~\cite{he2016deep} without pooling layer after the input layer. In the transition stage, each Bi-directional Feature Integration (BFI) block collects two or three adjacent encoder features to generate new feature $b_k$.  
	In the refinement stage, each Mask-guided Dual Attention (MDA) block takes $b_k$ as input,  refines them via channel attention and spatial attention guided by inharmonious region mask, and produces the refined encoder feature $a_k$.
	In the decoding stage, refined encoder features $\{a_k|_{k=1}^5\}$ are fed into Global-context Guided Decoder (GGD) blocks to predict the inharmonious region mask  $\hat{M}$, during which the global-context feature is injected into each GGD via a shortcut. Next, we will introduce our BFI block, MDA block, and GGD block one by one.
	
	\subsection{Bi-directional Feature Integration (BFI) Block} \label{sec:BFI}
	The encoder features $\{r_k|_{k=1}^5\}$ from ResNet encoder blocks contain the contextual information of different scales. As discussed in Section~\ref{sec:intro}, observing multi-scale contextual information could make it easier to locate the inharmonious region, so we attempt to fuse multi-scale features in the transition stage. 
	Inspired by  Aggregate Interaction Module (AIM) in \cite{pang2020multi}, we aggregate the adjacent encoder features because adjacent ones are more relevant without huge resolution gap. According to the position of BFI block (leftmost, middle, rightmost), we design three types of BFI blocks as shown in the right of Fig.~\ref{fig:network}.
	
	The structure of middle BFI block is shown in BFI(a) in the right of Fig.~\ref{fig:network}. Each BFI block has three streams: top-down stream (first row), bottom-up stream (second row), and aggregation stream (third row). We use $r_{k+1}$, $r_{k}$, $r_{k-1}$ to denote three encoder features from high-level to low-level. 
	Top-down (\emph{resp.}, bottom-up) stream propagates information from high-level (\emph{resp.}, low-level) to low-level (\emph{resp.}, high-level).
	In the top-down stream, high-level encoder feature $b^{\downarrow}_{k+1}=r_{k+1}$ is passed through an upsampling layer $U$ (upsampling operation followed by a $3\times 3$ convolution layer with stride=1) and added to $r_k$, resulting in $b^{\downarrow}_{k}=U(b^{\downarrow}_{k+1})+r_k$. $b^{\downarrow}_{k}$ is again passed through $U$ and added to $r_{k-1}$, resulting in $b^{\downarrow}_{k-1}=U(b^{\downarrow}_{k})+r_{k-1}$. Similarly, in the bottom-up stream, we can obtain $b^{\uparrow}_{k-1}$, $b^{\uparrow}_{k}$, $b^{\uparrow}_{k+1}$.
	In the aggregation stream, we aggregate the information from both top-down stream and bottom-up stream. Specifically, we concatenate $b^{\downarrow}_{k-1}$ and $b^{\uparrow}_{k-1}$, and pass them through a downsampling layer $D$ (a $3\times 3$ convolution layer with stride=2), leading to $D([b^{\downarrow}_{k-1}, b^{\uparrow}_{k-1}])$. Similarly, we concatenate $b^{\downarrow}_{k+1}$ and $b^{\uparrow}_{k+1}$, and pass them through an upsampling layer $U$, leading to $U([b^{\downarrow}_{k+1}, b^{\uparrow}_{k+1}])$. Finally, we sum up all the features with the same resolution as $r_{k}$, yielding the output of BFI block: $b_k = D([b^{\downarrow}_{k-1}, b^{\uparrow}_{k-1}])+U([b^{\downarrow}_{k+1}, b^{\uparrow}_{k+1}])+b^{\downarrow}_k+b^{\uparrow}_k$.

	The other two types of BFI blocks (see BFI(b)(c) in the right of Fig.~\ref{fig:network}) are designed in a similar way and we omit the details here. Compared with AIM~\cite{pang2020multi}, our BFI enhances multi-scale features via bi-directional information flow and  integrates multi-scale features more thoroughly (see supplementary for detailed comparison).

	\subsection{Mask-guided Dual Attenton (MDA) Block}
	In the transition stage, we combine the features of different scales and obtain the rich context-aware features. However, multi-scale feature fusion will inevitably introduce redundant information~\cite{pang2020multi,zhang2018bi}, that is, not all outputs contribute to the final inharmonious region prediction. The related inharmonious object clues should be retained and the other distractors should be abandoned. To this end, we employ Mask-guided Dual Attention (MDA) block to suppress the redundant information in the refinement stage. Our MDA block is built upon Convolutional Block Attention Module (CBAM)~\cite{woo2018cbam}. CBAM is composed of dual attention, \emph{i.e.}, channel attention and spatial attention, which can attend the informative contexts in the region of interests. For the details of channel and spatial attention, please refer to \cite{woo2018cbam}.  
	
	To make the network focus on inharmonious region and alleviate the attention drift problem~\cite{cheng2017focusing}, we add supervision to the spatial attention map, that is, enforcing the learnt spatial attention map to approach the ground-truth inharmonious region mask. We refer to the output from the $k$-th MDA block as refined encoder feature $a_k$.

	\vspace{-0.2cm}
	\subsection{Global-context Guided Decoder (GGD) Block}
	In the decoding stage, we process the refined encoder features $\{a_k |_{k=1}^5\}$ in a cascaded way. As shown in the right of Fig.~\ref{fig:network}, our GGD block has two streams. The top stream is similar to the skip connection in UNet-like network structure. In the $k$-th GGD block, we pass the output from the $(k+1)$-th GGD block through an upsampling layer (the same as $U$ in Section~\ref{sec:BFI}) and multiply it with $a_k$. We choose multiplication here, because  multiplication can reduce the gap between multi-level features~\cite{wu2019cascaded} and we find it empirically effective in the decoder (see supplementary). %The operator $\text{BConv}$ includes a sequence of operations $\text{BatchNorm}(\text{ReLU}(\text{Conv}(x)))$. 
	
	Besides, we realize that global contextual information (\emph{e.g.}, scene type, global lighting condition) can provide useful guidance for identifying the inharmonious region. As the high-level refined encoder feature $a_5$ encodes the global contextual information, we add a shortcut in the bottom stream to inject $a_5$ or upsampled $a_5$ into each GGD block, and concatenate with the output of top stream. In this way, GGD is equipped with sufficient guidance from the perspective of global context. 
	Finally, the top stream of the lowest-level GGD block will output the inharmonious region mask $\hat{M}$.
	
	\subsection{Loss Function}
	Let $M$ be the ground-truth inharmonious region mask, $\hat{M}$ be the predicted mask, $A_{k}$ be the learnt spatial attention map from the $k$-th MDA block. We calculate three losses: binary cross-entropy loss $\mathcal{L}_{bce}$, structural similarity loss $\mathcal{L}_{ssim}$, and auxiliary attention loss $\mathcal{L}_{aux}$.  $\mathcal{L}_{bce}$ is widely used in the binary classification task and segmentation task:
	\begin{eqnarray}
	\mathcal{L}_{bce} \!=\! -\!\sum_{i,j}{M}_{i,j}\text{log}(\hat{M}_{i,j}) \!-\!\sum_{i,j}(1\!-\!M_{i,j})\text{log}(1\!-\!\hat{M}_{i,j}), \nonumber
	\end{eqnarray}
	in which the subscript $i,j$ indicates the pixel location.
	$\mathcal{L}_{ssim}$ is recommended by \cite{Qin_2019_CVPR} to better capture the structural information:
	\begin{eqnarray}
	\mathcal{L}_{ssim} = 1-\frac{(2\mu_x\mu_y + C_1)(2\sigma_{xy}+C_2)}{(\mu^2_x +  \mu^2_y+C_1)(\sigma^2_{x} +     \sigma^2_{y}+C_2)},
	\end{eqnarray}
	in which $x,y$ represent the patches cropped from the predicted mask and the ground-truth mask respectively. $\mu_x, \mu_y, \sigma_x, \sigma_y$, $\sigma_{xy}$ are the mean of  $x,y$, standard deviation of $x,y$, covariance between $x,y$, respectively. $C_1$ and $C_2$ are two constants which are set to $0.01^2$ and $0.03^2$ respectively according to \cite{Qin_2019_CVPR}.
	$\mathcal{L}_{aux}$ shares the similar form as $\mathcal{L}_{bce}$ and guides the learning of spatial attention map:
	\begin{eqnarray}
	\mathcal{L}_{aux} \!=\!\!\!\!\!\!\!&&-\sum_{k}\sum_{i,j}{M}_{i,j}\text{log}(A_{k,i,j}) \nonumber \\ 
	&&-\sum_k\sum_{i,j}(1-M_{i,j})\text{log}(1-A_{k,i,j}),
	\end{eqnarray}
	in which $A_{k,i,j}$ is the $(i,j)$-th entry in $A_k$. The ground-truth mask $M$ is resized to match the resolution of each $A_k$.
	The total loss can be written as
	\begin{eqnarray}
	\mathcal{L}_{total} = \mathcal{L}_{bce} + \mathcal{L}_{ssim} + \lambda \mathcal{L}_{aux},
	\end{eqnarray}
	where the trade-off parameter $\lambda$ is empirically set as 0.1.
	
	\begin{figure*}[!ht]
		\centering
		\includegraphics[scale=0.6]{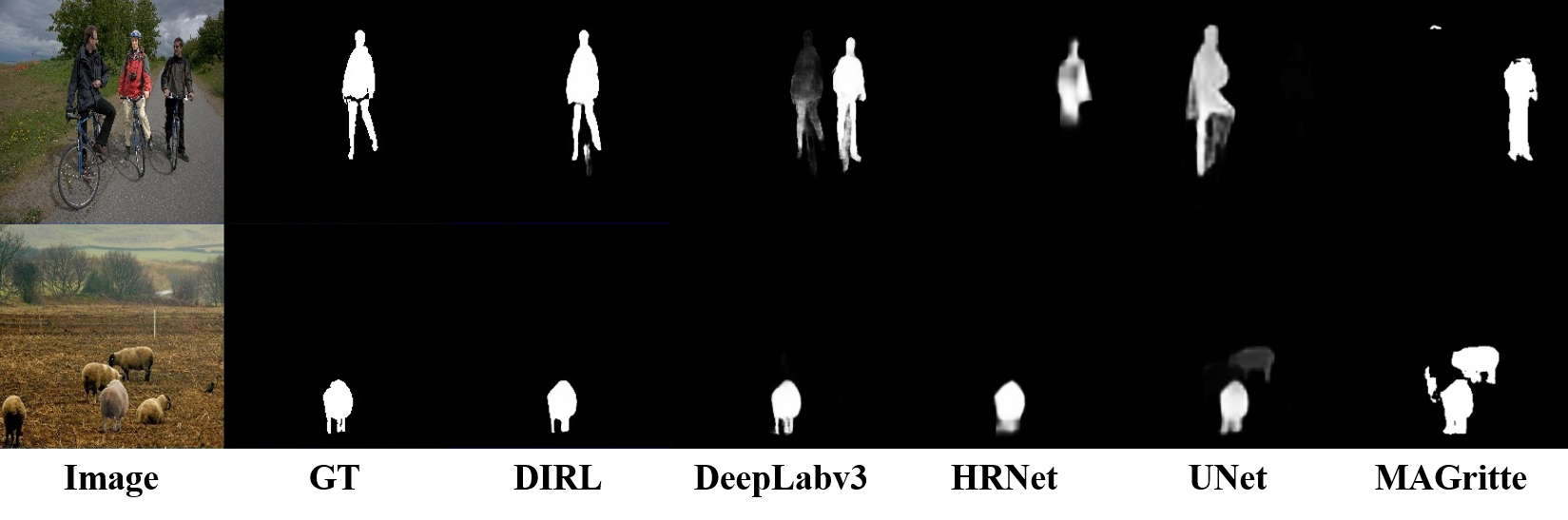}
		\caption{Comparison with the state-of-the-art methods from related fields. Our DIRL can localize the inharmonious region around the complex background and accurately capture the outline of inharmonious area. }
		\label{fig:mask1}
	\end{figure*}
	
	\section{Experiments}
	
	\vspace{-0.2cm}
	\subsection{Experimental Setting}
	We conduct experiments on the image harmonization dataset \textbf{iHarmony4}~\cite{cong2019deep}, which provides pairs of inharmonious and harmonious images. iHarmony4 is composed of four sub-datasets: HAdobe5K, HCOCO, HFlickr, HDay2Night. For HCOCO and HFlickr datasets, the inharmonious images are obtained by adjusting the color and lighting statistics of foreground. For HAdobe5K and HDay2Night datasets, the inharmonious images are obtained by overlaying the foreground with the counterpart of the same scene retouched with a different style or captured in a different condition. For the inharmonious images in all four sub-datasets, the foreground region looks incompatible with the background mainly due to color and lighting inconsistency, which enables us to focus on inharmonious region localization by using this dataset. 
	In this work, we only use the inharmonious images without using paired harmonious images. One concern is that the inharmonious region in an inharmonious image may be ambiguous because the background can also be treated as inharmonious region. 
	To avoid the ambiguity, we simply discard the images with foreground occupying larger than 50\% area, which only account for about 2\% of the whole dataset. This strategy is similar to previous image manipulation localization works~\cite{Wu_2019_CVPR} and our task becomes localizing the inharmonious region which occupies smaller than 50\% area.  %huh18forensics,
	Considering the data split in~\cite{cong2019deep} and the above concern, we tailor the training set to 64255 images and test set to 7237 images respectively. 
	For quantitative evaluation, we calculate Average Precision (AP), $F_1$ score, and Intersection over Union (IoU) based on the predicted mask $\hat{M}$ and the ground-truth mask $M$. The details of evaluation metrics and our implementation are left to supplementary. 
	
	\vspace{-0.2cm}
	\subsection{Comparison with the State-of-the-art}
	Since this is the first work focusing on inharmonious region localization, we compare with the state-of-the-art methods from related fields. In particular, we compare with three groups of baselines: 1) image manipulation localization methods MFCN~\cite{Salloum_2018}, MantraNet~\cite{Wu_2019_CVPR}, MAGritte~\cite{kniaz2019point}, H-LSTM~\cite{bappy2019hybrid}; 2) image harmonization method S2AM~\cite{cun2020improving}, which can predict the inharmonious region mask as a byproduct; 3) popular segmentation networks UNet~\cite{ronneberger2015u}, Deeplabv3~\cite{chen2017rethinking}, and HRNet~\cite{wang2020deep}. For the first two groups, we conduct the experiments based on their released code. For the third group, we use ResNet34 as backbone for UNet and Deeplabv3. For HRNet, we use HRNet30 since its model size is comparable with ResNet34. 
	%Moreover, all of the results are generated by the model prediction without any post-processing and assessed by the same evaluation code.
	
	\textbf{Quantitative Comparisons:} The results of different methods are summarized in Table~\ref{tab:baselines}, we can observe that our DIRL  beats the best method MAGritte~\cite{kniaz2019point} from the field of image manipulation localization and  S2AM~\cite{cun2020improving} from the field of image harmonization by a large margin. Another observation is that typical segmentation methods achieve competitive performance, even though they are not specifically designed for inharmonious region localization. However, our DIRL still achieves 4.33\% (\emph{resp.}, 4.15\%, 1.84\%) improvement for AP, $F_1$, and IoU against the strongest baseline, which demonstrates the effectiveness of our proposed method.
	
	\textbf{Qualitative Comparisons:}
	We also show the predicted masks of different methods in Fig.~\ref{fig:mask1}. 
	%The images are selected from the four sub-datasets from iHarmoney4. 
	%It is obvious that our method can  detect the inharmonious regions more accurately. 
	For some challenging scenarios, our DIRL can successfully identify the inharmonious region while other methods (\emph{e.g.}, UNet and MAGritte) are struggling to identify the inharmonious region (see the first row). Besides, compared with baselines (\emph{e.g.}, HRNet, UNet), DIRL can localize the intact inharmonious region with sharp boundary (see the second row). More qualitative comparisons could be found in the supplementary.

	\vspace{-0.2cm}
	\subsection{Ablation Studies}
	In this section, we conduct ablation studies to verify the effectiveness of our BFI block, MDA block, and GGD block. We also visualize the spatial attention map learnt by MDA block. The details are left to supplementary due to space limitation.
	
	\begin{table}[t]
		\caption{AP, $F_1$, IoU of different methods on the iHarmony4 dataset. $\uparrow$ means the larger, the better. The best results are denoted in boldface.}
		\centering
		\label{tab:baselines}
		\setlength{\tabcolsep}{3mm}{
			\begin{tabular}{l c c c}
				\toprule[1pt]
				% 		\vspace{-2pt}
				{Method}  & AP(\%) $\uparrow$ & $F_1 \uparrow$ & IoU(\%) $\uparrow$ \cr \hline \hline
				\text{UNet}~\cite{ronneberger2015u} & 74.90 & 0.6717 & 64.74 \\
				\text{DeepLabv3}~\cite{chen2017rethinking} & 75.69  & 0.6902 & 66.01 \\
				\text{HRNet}~\cite{wang2020deep} & 75.33 & 0.6765 & 65.49 \\
				\hline
				\text{MFCN}~\cite{Salloum_2018} & 45.63 & 0.3794 & 28.54 \\
				\text{MantraNet}~\cite{Wu_2019_CVPR} &  64.22 & 0.5691 & 50.31 \\
				\text{MAGritte}~\cite{kniaz2019point} &  71.16 & 0.6907  & 60.14 \\
				\text{H-LSTM}~\cite{bappy2019hybrid} &  60.21 & 0.5239 & 47.07 \\
				\hline
				\text{S2AM}~\cite{cun2020improving} &  43.77 & 0.3029 & 22.36 \\
				\hline
				\textbf{DIRL} & \textbf{80.02} & \textbf{0.7317} & \textbf{67.85} \\
				\bottomrule[1pt]
			\end{tabular}
		}
	\end{table}
 
	\vspace{-0.2cm}
	\section{Conclusion}
	In this paper, we are the first to focus on inharmonious region localization task. We have proposed a novel network including BFI block in the transition stage, MDA block in the refinement stage, and GGD block in the decoding stage, which can effectively fuse multi-scale context and suppress redundancy. Experiments on iHarmony4 have demonstrated the superiority of our proposed method.
	
	\vspace{-0.3cm}
	\section{Acknowledgement}
	The work is supported by the National Key R\&D Program of China (2018AAA0100704) and is partially sponsored by National Natural Science Foundation of China (Grant No.61902247) and Shanghai Sailing Program (19YF1424400).
	
	\vspace{-0.2cm}
	\bibliographystyle{IEEEbib}
	\bibliography{main}

\begin{thebibliography}{10}

\bibitem{tsai2017deep}
Yi-Hsuan Tsai, Xiaohui Shen, Zhe Lin, Kalyan Sunkavalli, Xin Lu, and Ming-Hsuan
  Yang,
\newblock ``Deep image harmonization,''
\newblock in {\em CVPR}, 2017.

\bibitem{cun2020improving}
Xiaodong Cun and Chi-Man Pun,
\newblock ``Improving the harmony of the composite image by spatial-separated
  attention module,''
\newblock {\em TIP}, 2020.

\bibitem{cong2019deep}
Wenyan Cong, Jianfu Zhang, Li~Niu, Liu Liu, Zhixin Ling, Weiyuan Li, and Liqing
  Zhang,
\newblock ``Dovenet: Deep image harmonization via domain verification,''
\newblock {\em CVPR}, 2020.

\bibitem{pang2020multi}
Youwei Pang, Xiaoqi Zhao, Lihe Zhang, and Huchuan Lu,
\newblock ``Multi-scale interactive network for salient object detection,''
\newblock in {\em CVPR}, 2020.

\bibitem{zhang2018bi}
Lu~Zhang, Ju~Dai, Huchuan Lu, You He, and Gang Wang,
\newblock ``A bi-directional message passing model for salient object
  detection,''
\newblock in {\em CVPR}, 2018.

\bibitem{gu2020hard}
Zhangxuan Gu, Li~Niu, Haohua Zhao, and Liqing Zhang,
\newblock ``Hard pixel mining for depth privileged semantic segmentation,''
\newblock {\em T-MM}, 2020.

\bibitem{zhang2017amulet}
Pingping Zhang, Dong Wang, Huchuan Lu, Hongyu Wang, and Xiang Ruan,
\newblock ``Amulet: Aggregating multi-level convolutional features for salient
  object detection,''
\newblock in {\em ICCV}, 2017.

\bibitem{isola2017image}
Phillip Isola, Jun-Yan Zhu, Tinghui Zhou, and Alexei~A Efros,
\newblock ``Image-to-image translation with conditional adversarial networks,''
\newblock in {\em CVPR}, 2017.

\bibitem{woo2018cbam}
Sanghyun Woo, Jongchan Park, Joon-Young Lee, and In~So~Kweon,
\newblock ``{CBAM}: Convolutional block attention module,''
\newblock in {\em ECCV}, 2018.

\bibitem{cheng2017focusing}
Zhanzhan Cheng, Fan Bai, Yunlu Xu, Gang Zheng, Shiliang Pu, and Shuigeng Zhou,
\newblock ``Focusing attention: Towards accurate text recognition in natural
  images,''
\newblock in {\em ICCV}, 2017.

\bibitem{chen2017image}
Can Chen, Scott McCloskey, and Jingyi Yu,
\newblock ``Image splicing detection via camera response function analysis,''
\newblock in {\em CVPR}, 2017.

\bibitem{hu2016effective}
Wu-Chih Hu, Wei-Hao Chen, Deng-Yuan Huang, and Ching-Yu Yang,
\newblock ``Effective image forgery detection of tampered foreground or
  background image based on image watermarking and alpha mattes,''
\newblock {\em Multimed. Tools. Appl.}, 2016.

\bibitem{ZHAO2013158}
Jie Zhao and Jichang Guo,
\newblock ``Passive forensics for copy-move image forgery using a method based
  on {DCT} and {SVD},''
\newblock {\em Forensic Sci. Int.}, 2013.

\bibitem{6151134}
Tiziano Bianchi and Alessandro Piva,
\newblock ``Image forgery localization via block-grained analysis of {JPEG}
  artifacts,''
\newblock {\em IEEE Trans. Inf. Forensics Secur.}, 2012.

\bibitem{Wu_2019_CVPR}
Yue Wu, Wael AbdAlmageed, and Premkumar Natarajan,
\newblock ``Mantra-net: Manipulation tracing network for detection and
  localization of image forgeries with anomalous features,''
\newblock in {\em CVPR}, 2019.

\bibitem{bappy2019hybrid}
Jawadul~H Bappy, Cody Simons, Lakshmanan Nataraj, BS~Manjunath, and Amit~K
  Roy-Chowdhury,
\newblock ``Hybrid {LSTM} and encoder-decoder architecture for detection of
  image forgeries,''
\newblock {\em TIP}, 2019.

\bibitem{bahrami2015blurred}
Khosro Bahrami, Alex~C Kot, Leida Li, and Haoliang Li,
\newblock ``Blurred image splicing localization by exposing blur type
  inconsistency,''
\newblock {\em IEEE Trans. Inf. Forensics Secur.}, 2015.

\bibitem{bappy2017exploiting}
Jawadul~H Bappy, Amit~K Roy-Chowdhury, Jason Bunk, Lakshmanan Nataraj, and
  BS~Manjunath,
\newblock ``Exploiting spatial structure for localizing manipulated image
  regions,''
\newblock in {\em ICCV}, 2017.

\bibitem{Salloum_2018}
Ronald Salloum, Yuzhuo Ren, and C.-C. Jay~Kuo,
\newblock ``Image splicing localization using a multi-task fully convolutional
  network ({MFCN}),''
\newblock {\em J. Vis. Commun. Image Represent.}, 2018.

\bibitem{kniaz2019point}
Vladimir~V Kniaz, Vladimir Knyaz, and Fabio Remondino,
\newblock ``The point where reality meets fantasy: Mixed adversarial generators
  for image splice detection,''
\newblock in {\em NeurIPS}, 2019.

\bibitem{9102825}
C.~{Yang}, H.~{Li}, F.~{Lin}, B.~{Jiang}, and H.~{Zhao},
\newblock ``Constrained {R-CNN}: A general image manipulation detection
  model,''
\newblock in {\em ICME}, 2020.

\bibitem{lin2017feature}
Tsung-Yi Lin, Piotr Doll{\'a}r, Ross Girshick, Kaiming He, Bharath Hariharan,
  and Serge Belongie,
\newblock ``Feature pyramid networks for object detection,''
\newblock in {\em CVPR}, 2017.

\bibitem{liu2018path}
Shu Liu, Lu~Qi, Haifang Qin, Jianping Shi, and Jiaya Jia,
\newblock ``Path aggregation network for instance segmentation,''
\newblock in {\em CVPR}, 2018.

\bibitem{gu2020context}
Zhangxuan Gu, Siyuan Zhou, Li~Niu, Zihan Zhao, and Liqing Zhang,
\newblock ``Context-aware feature generation for zero-shot semantic
  segmentation,''
\newblock in {\em ACM MM}, 2020.

\bibitem{ronneberger2015u}
Olaf Ronneberger, Philipp Fischer, and Thomas Brox,
\newblock ``U-net: Convolutional networks for biomedical image segmentation,''
\newblock in {\em MICCAI}, 2015.

\bibitem{he2016deep}
Kaiming He, Xiangyu Zhang, Shaoqing Ren, and Jian Sun,
\newblock ``Deep residual learning for image recognition,''
\newblock in {\em CVPR}, 2016.

\bibitem{wu2019cascaded}
Zhe Wu, Li~Su, and Qingming Huang,
\newblock ``Cascaded partial decoder for fast and accurate salient object
  detection,''
\newblock in {\em CVPR}, 2019.

\bibitem{Qin_2019_CVPR}
Xuebin Qin, Zichen Zhang, Chenyang Huang, Chao Gao, Masood Dehghan, and Martin
  Jagersand,
\newblock ``Basnet: Boundary-aware salient object detection,''
\newblock in {\em CVPR}, 2019.

\bibitem{chen2017rethinking}
Liang-Chieh Chen, George Papandreou, Florian Schroff, and Hartwig Adam,
\newblock ``Rethinking atrous convolution for semantic image segmentation,''
\newblock {\em arXiv preprint arXiv:1706.05587}, 2017.

\bibitem{wang2020deep}
Jingdong Wang, Ke~Sun, Tianheng Cheng, Borui Jiang, Chaorui Deng, Yang Zhao,
  Dong Liu, Yadong Mu, Mingkui Tan, Xinggang Wang, et~al.,
\newblock ``Deep high-resolution representation learning for visual
  recognition,''
\newblock {\em TPAMI}, 2020.

\end{thebibliography}


\begin{thebibliography}{1}

\bibitem{pang2020multi}
Youwei Pang, Xiaoqi Zhao, Lihe Zhang, and Huchuan Lu,
\newblock ``Multi-scale interactive network for salient object detection,''
\newblock in {\em CVPR}, 2020.

\bibitem{paszke2019pytorch}
Adam Paszke, Sam Gross, Francisco Massa, Adam Lerer, James Bradbury, Gregory
  Chanan, Trevor Killeen, Zeming Lin, Natalia Gimelshein, Luca Antiga, et~al.,
\newblock ``Pytorch: An imperative style, high-performance deep learning
  library,''
\newblock in {\em NeurIPS}, 2019.

\bibitem{he2016deep}
Kaiming He, Xiangyu Zhang, Shaoqing Ren, and Jian Sun,
\newblock ``Deep residual learning for image recognition,''
\newblock in {\em CVPR}, 2016.

\bibitem{cong2019deep}
Wenyan Cong, Jianfu Zhang, Li~Niu, Liu Liu, Zhixin Ling, Weiyuan Li, and Liqing
  Zhang,
\newblock ``Dovenet: Deep image harmonization via domain verification,''
\newblock {\em CVPR}, 2020.

\bibitem{wu2019cascaded}
Zhe Wu, Li~Su, and Qingming Huang,
\newblock ``Cascaded partial decoder for fast and accurate salient object
  detection,''
\newblock in {\em CVPR}, 2019.

\bibitem{cheng2017focusing}
Zhanzhan Cheng, Fan Bai, Yunlu Xu, Gang Zheng, Shiliang Pu, and Shuigeng Zhou,
\newblock ``Focusing attention: Towards accurate text recognition in natural
  images,''
\newblock in {\em ICCV}, 2017.

\end{thebibliography}
	
\end{document}

\sloppy

% Example definitions.
% --------------------
\def\x{{\mathbf x}}
\def\L{{\cal L}}
\title{Supplementary for Inharmonious Region Localization}

\name{Jing Liang, Li Niu\textsuperscript{*}\thanks{\textsuperscript{*}Corresponding author.},  Liqing Zhang}
\address {MoE Key Lab of Artificial Intelligence, Department of Computer Science and Engineering \\
Shanghai Jiao Tong University, Shanghai, China \\ 
\{leungjing, ustcnewly\}@sjtu.edu.cn, zhang-lq@cs.sjtu.edu.cn.}
\maketitle

In this supplementary, we will provide the details of our implementation in Section~\ref{sec:implementation} and the details of evaluation metrics in Section~\ref{sec:evaluation}. We will compare our BFI block and AIM block~\cite{pang2020multi} in Section~\ref{sec:aim_bfi}, and perform ablation studies on our proposed blocks in Section~\ref{sec:ablate}. We will show the visualization results of spatial attention maps in Section~\ref{sec:MDA} and predicted masks in Section~\ref{sec:visualization}.

\section{Implementation Details}\label{sec:implementation}

We implement our model using Pytorch~\cite{paszke2019pytorch} and use Adam optimizer with $\beta_1=0.9$, $\beta_2=0.999$, weight decay being 1e-4, initial learning rate being 1e-4. We apply multi-step decay strategy to reduce the learning rate after training 30 epochs with a factor of 0.5. 

We use five res-blocks as our encoder, where the first four are adopted from ResNet34~\cite{he2016deep} and the last one is similar to the first block in ResNet34. Besides, we replace the input $7 \times 7$ convolution layer with a $3 \times 3$ convolution layer and remove pooling layer to better keep the details in the shallow layers. 

\section{Evaluation Metrics}\label{sec:evaluation}
In our experiments, the ground-truth mask $M$ and the predicted mask $\hat{M}$
are used to calculate the $Precision = \frac{TP}{ (TP + FP)}$ and $Recall = \frac{TP}{(TP+FN)}$, where $TP, FP$, and $FN$ represent true-positive, false-positive and false-negative, respectively. Based on $TP$, $FP$, $FN$, we calculate $AP$, $F_1$, and $IoU$ for each image, and then calculate the average over all test images. 

For Average Precision ($AP$), we obtain the average precision by accumulating the precision at different
thresholds:
\begin{eqnarray}
AP = \sum_{n=0}^{255} ({Recall}_{n+1} - {Recall}_{n})\times {Precision}_{n},
\end{eqnarray}
where ${Recall}_n$ and ${Precision}_n$ are the precision and recall at the $n$-th threshold respectively.

The F-measure is an overall performance indicator, which is computed by the weighted harmonic of precision and recall: 
\begin{eqnarray}
F_{\beta} = \frac{(1+\beta^2){Precision} \times {Recall}}{\beta^2 {Precision} + {Recall}},
\end{eqnarray}
where we use $0.5$ as the threshold and set $\beta = 1$, resulting in $F_1$-score.

We also take the Intersection over Union ($IoU$) metric into account as it reflects the similarity between the predicted mask and the ground-truth mask:
\begin{eqnarray}
IoU = \frac{TP}{TP+FP+FN},
\end{eqnarray}
in which we also use $0.5$ as the threshold.

\begin{figure*}[!ht]
\centering
\includegraphics[scale=0.5]{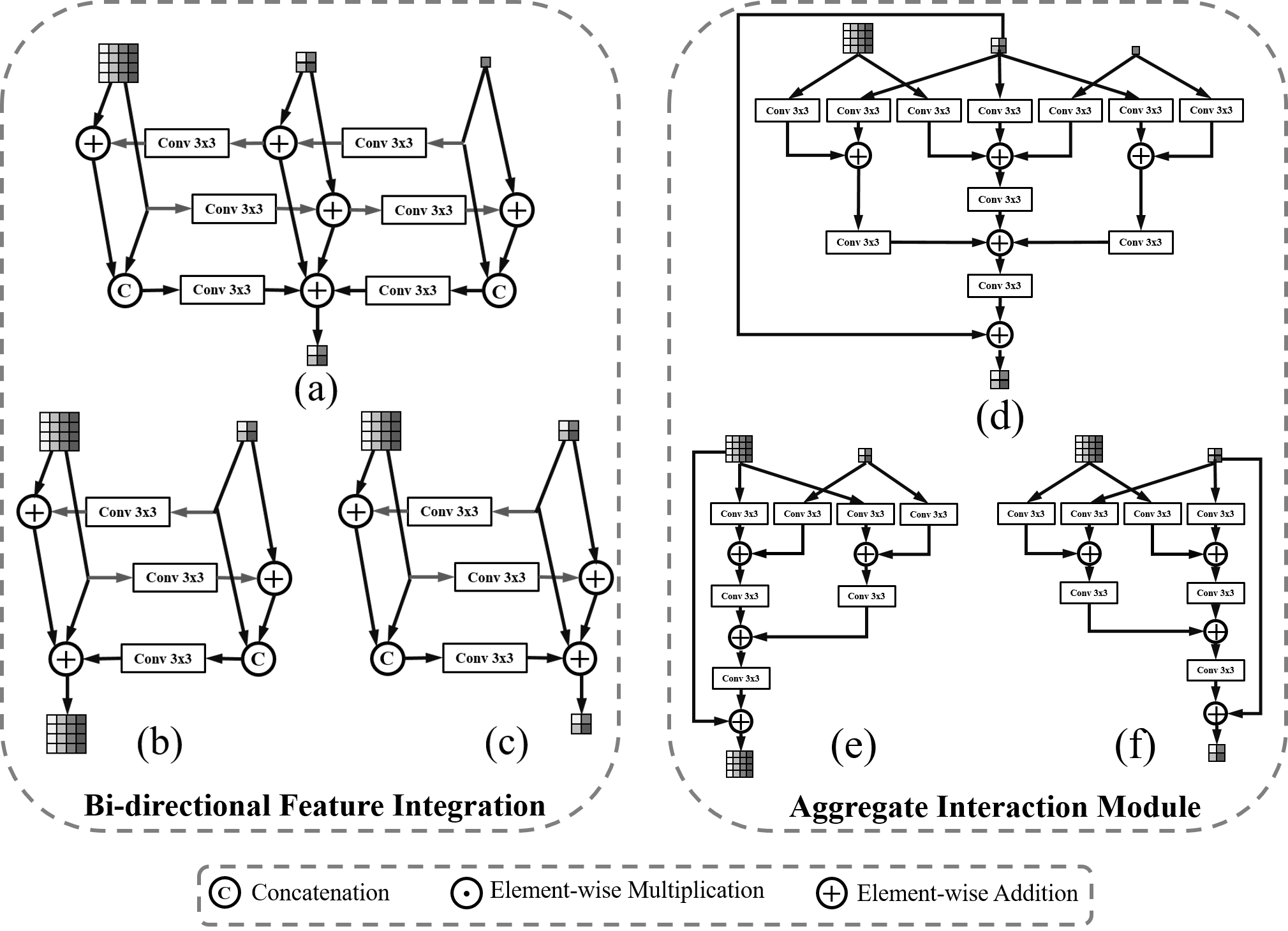}
\caption{The left part exhibits three types of Bi-directional Feature Integration (BFI) block. The right part exhibits three types of Aggregate Integration Module (AIM). ``Conv k x k" means the convolution layer with kernel size $k$.}
\label{fig:integration}
\end{figure*}

\section{Comparison between AIM and our BFI block} \label{sec:aim_bfi}
Our Bidirectional Feature Integration (BFI) block is similar to Aggregate Interaction Module (AIM) in \cite{pang2020multi}, which also fuses two or three adjacent encoder features. We compare the network structure of BFI and AIM in Fig.~\ref{fig:integration}. According to the block position in the transition stage, both BFI and AIM  fall into three types: leftmost, middle, rightmost. For BFI, the structure of leftmost (\emph{resp.}, middle, rightmost) block is shown in  (b) (\emph{resp.}, (a), (c)). For AIM, the structure of leftmost (\emph{resp.}, middle, rightmost) block is shown in  (e) (\emph{resp.}, (d), (f)). Next, we compare the AIM middle block and BFI middle block. The other two types of blocks can be compared similarly.

To formulate the aggregation procedure of AIM, we follow the description and symbols used for BFI. Let $U$ be the upsampling layer (upsampling operator followed by a $3 \times 3$ convolution layer with stride one) and $D$ be the downsampling layer (a $3 \times 3$ convolution layer with stride two). A regular convolution layer is dubbed as $Conv$ and the $k$-th encoder feature is $r_k$. The output of the $k$-th AIM block $b_k$ is obtained as follows:
\begin{eqnarray}
\begin{aligned}
    z_{k-1} &=& & Conv(U(r_k) + Conv(r_{k-1})), \\
    z_k &=&  &    Conv(D(r_{k-1}) + Conv(r_k) + U(r_{k+1})), \\
    z_{k+1} &=& & Conv(D(r_k) + Conv(r_{k+1})), \\ 
    b_k &=& & Conv(z_{k-1} + z_k + z_{k+1})  + r_k.
\end{aligned}
\end{eqnarray}

\begin{table*}[ht]
\centering
\caption{Ablation studies on our BFI block, MDA block, and GGD block on iHarmony4 dataset. BFI($\uparrow$) (\emph{resp.}, BFI($\downarrow$)) is a special case of BFI which only contains the bottom-up (\emph{resp.}, top-down) stream. DA means the Dual Attention block without mask supervision. Reg means regular decoder used in UNet. GGD\_sim means GGD without global-context shortcut.}
\label{tab:ablation}
\vspace{3 pt}
% \resizebox{1\linewidth}{!}
\setlength{\tabcolsep}{7mm}
  {
  \begin{tabular} {c | c | c | c | c c c }
    \toprule[1pt]
 \multirow{2}{*}{\textbf{\#}} & \multirow{2}{*}{\textbf{Transition}} & \multirow{2}{*}{\textbf{Refine}} & \multirow{2}{*}{\textbf{Decoder}} & \multicolumn{3}{c}{\textbf{Evaluation Metrics}} \\ \cline{5-7}
    & & & &\multicolumn{1}{c|}{AP(\%)} &  \multicolumn{1}{c|}{$F_1$} & \multicolumn{1}{c}{IoU(\%)} 
    \\    \hline \hline
    1 &-&- & Reg     & 74.90 &0.6717 & 64.74 \\
    2 &- &- & GGD\_sim & 76.91 & 0.7055 & 65.31\\
    3 &- & -& GGD & 77.99  & 0.7131 & 66.19 \\
    4 & AIM~\cite{pang2020multi} &- & GGD &  77.86   & 0.7116  &  65.94 \\
    5 &  BFI($\downarrow$) & -& GGD &    78.33     &  0.7190 & 66.64  \\
    6 & BFI($\uparrow$) & -& GGD &   78.06     &  0.7127 & 66.49  \\
    7 &  BFI &- & GGD &   78.56   & 0.7205 & 66.85   \\
    8 & -   & DA & GGD &  79.28  &0.7258 & 67.16  \\
    9 & -   & MDA & GGD &    79.38   & 0.7311 & 67.65  \\
    10 &  BFI   & MDA & GGD &    80.02   & 0.7317 & 67.85 \\
    \bottomrule[1pt]
  \end{tabular}
  }
\end{table*}

As described in the main paper, the output of the $k$-th BFI $b_k$ is obtained as follows:

\noindent
\begin{eqnarray}
\begin{aligned}
    &b_{k+1}^{\downarrow} &=&  r_{k+1}, &&b_{k+1}^{\uparrow} &=& r_{k+1} + D(b_{k}^{\uparrow})\\
    &b_{k}^{\downarrow} &=&  r_{k} + U(b_{k+1}^{\downarrow}), &&b_{k}^{\uparrow} &=& r_{k} + D(b_{k-1}^{\uparrow})\\ &b_{k-1}^{\downarrow} &=&  r_{k+1} + U(b_{k}^{\downarrow}),  &&b_{k-1}^{\uparrow} &=&  r_{k-1}
     \\
    &z_{k+1} &=&  U(b_{k+1}^{\downarrow}, b_{k+1}^{\uparrow}),&&z_k &=&  b_{k}^{\downarrow} + b_{k}^{\uparrow}, \\
    &z_{k-1} &=&  D(b_{k-1}^{\downarrow}, b_{k-1}^{\uparrow}),&&b_k &=&   z_{k-1} + z_k + z_{k+1}
    % , \\
    % , \\
    %  \\
    %  \\
    %  \\
    % .
\end{aligned}
\end{eqnarray}
where $\uparrow$ and $\downarrow$ refer to the bottom-up stream and top-down stream respectively.

It can be seen that AIM focuses on merging neighboring encoder features. There is no  information flow from the highest-level encoder feature to the lowest-level encoder feature, or the other way around. So the information exchange between the high-level encoder features and low-level encoder features is not sufficient. In contrast, 
our BFI accomplishes top-down (\emph{resp.}, bottom-up) information flow and obtain $b_i^{\downarrow}$ (\emph{resp.}, $b_i^{\uparrow}$) for $i\in\{k-1, k, k+1\}$. $b_i^{\uparrow}$ and $b_i^{\downarrow}$  are complementary to each other, which are used to obtain the transient features $z_{i}$ for each level $i\in\{k-1, k, k+1\}$. In this way, BFI can integrate multi-scale adjacent features more effectively and sufficiently.

\section{Ablation Studies}\label{sec:ablate}
Our proposed method consists of four stages: encoding stage, transition stage, refinement stage, and decoding stage. For the last three stages, we propose our Bi-directional Feature integration (BFI) block,  Mask-guided Dual Attention (MDA) block, and Global-context Guided Decoder (GGD) block, respectively. To investigate the effectiveness of each block, we conduct a series of experiments on iHarmony4~\cite{cong2019deep} dataset. We start from typical UNet and gradually build our method. 

\subsection{The Effectiveness of GGD}
Row 1 reports the performance of typical UNet without transition stage or refinement stage. We refer to the regular UNet-like connected decoder as ``Reg". We first replace ``Reg" with ``GGD\_sim". The only difference between ``Reg" and ``GGD\_sim" is whether using multiplication to integrate the encoder feature and the output from previous decoder block. By comparing row 1 and row 2, we find that multiplication performs more favorably, probably because  multiplication can reduce the gap between multi-level features~\cite{wu2019cascaded}.

Then we add the shortcut for global-context feature, leading to our GGD block. By comparing row 2 and row 3, it is useful to provide the guidance of global-context feature for each decoder block.

\subsection{The Effectiveness of BFI} 

Based on row 3, we investigate alternative blocks in the transition stage. We compare with AIM ~\cite{pang2020multi} discussed in Section~\ref{sec:aim_bfi}, which also fuses two or three adjacent encoder features. We also compare with two special cases of our BFI: BFI($\uparrow$) and BFI($\downarrow$).

For BFI($\uparrow$), we cut off the top-down stream and combine transient features $b_i^{\uparrow}$ for $i\in\{k-1, k, k+1\}$ as follows:
\begin{eqnarray}
b_k = D(b_{k-1}^{\uparrow}) + b_{k}^{\uparrow} + U(b_{k+1}^{\uparrow}).
\end{eqnarray}
Similarly, for BFI($\downarrow$), we cut off the bottom-up stream and combine transient features $b_i^{\downarrow}$ for $i\in\{k-1, k, k+1\}$ as follows:
\begin{eqnarray}
b_k = D(b_{k-1}^{\downarrow}) + b_{k}^{\downarrow} + U(b_{k+1}^{\downarrow}).
\end{eqnarray}

By comparing row 4$\sim$7, we observe that the performance of BFI will degenerate after discarding top-down stream or bottom-up stream. Our special cases BFI($\uparrow$) and BFI($\downarrow$) are slightly worse than or comparable with AIM. However, our intact BFI block achieves better results than AIM, which reveals the effectiveness of bi-directional information flow and aggregation.

\subsection{The Effectiveness of MDA}
To illustrate the effectiveness of our Mask-guided Dual Attention block, based on row 3, we add our Mask-guided Dual Attention (MDA) block and compare with our special case DA, which does not have mask supervision. The comparison between row 3 and row 8 demonstrates the effectiveness of using dual attention (spatial attention and channel attention) to suppress the redundancy information. The comparison between row 8 and row 9 proves that it is necessary to focus on inharmonious region and alleviate attention drift ~\cite{cheng2017focusing} by virtue of the supervision of ground-truth inharmonious region mask.

\begin{figure*}[!t]
\centering
\includegraphics[scale=0.55]{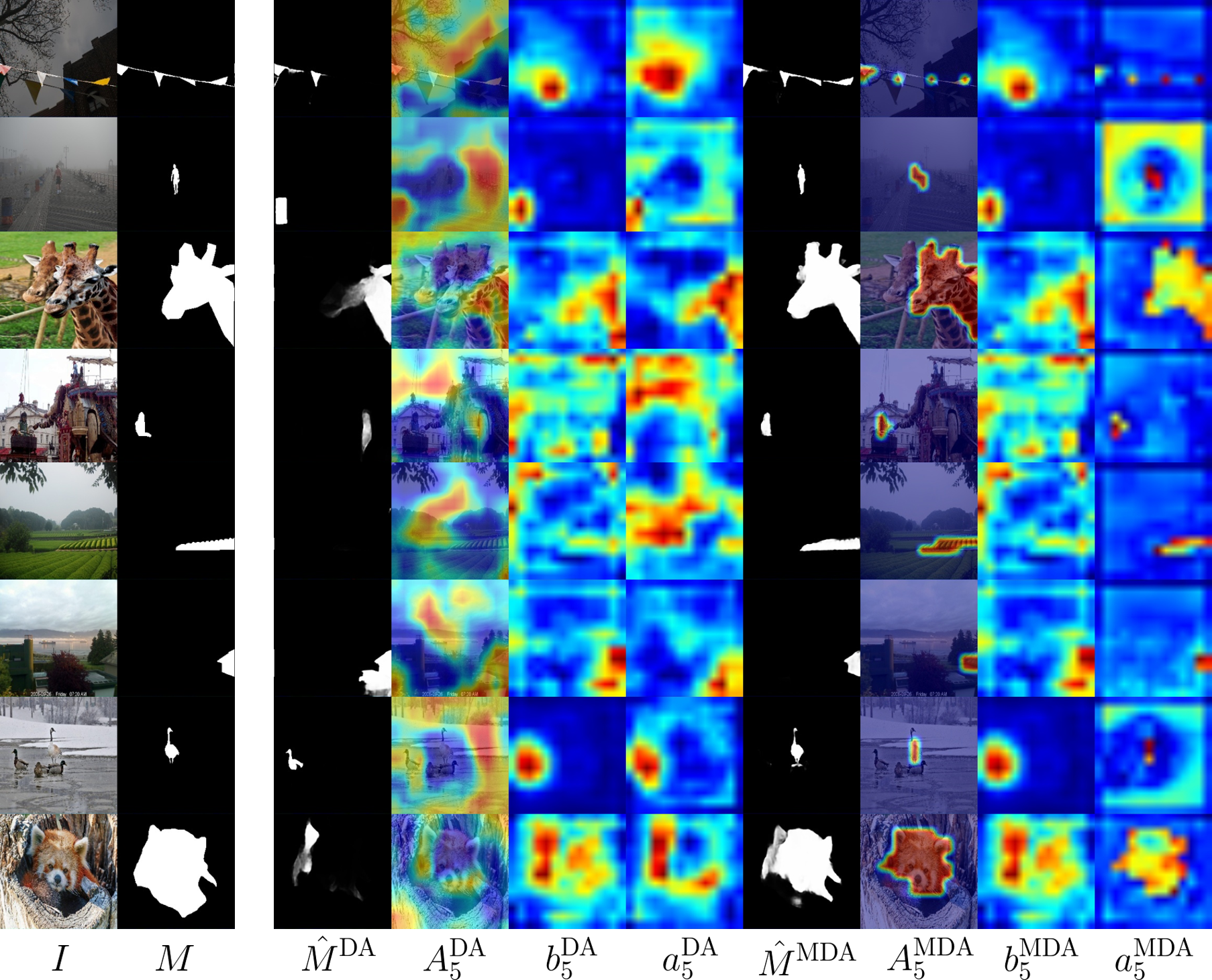}
\caption{Following the main paper, we use $I, M, \hat{M}, A_k, b_k, a_k$ to denote the input synthetic image, ground-truth mask, predicted mask, spatial attention map learnt by the $k$-th DA or MDA, the output feature of the $k$-th BFI, the output feature of the $k$-th DA or MDA, respectively. The superscript indicates that these results are obtained with Dual Attention (DA) block or Mask-guided Dual Attention (MDA) block.}
\label{fig:attention}
\end{figure*}

\section{Visualizing Spatial Attention Map in our MDA block} \label{sec:MDA}
We visualize the spatial attention maps learnt by DA without mask supervision and MDA with mask supervision  in Fig.~\ref{fig:attention}. Besides, to verify the importance of attention module in the refinement stage, we also exhibit the highest-level transition output feature $b_5$ and the refined encoder feature $a_5$ by means of heatmap.  By comparing the column $A_5^\text{DA}$ and the column $A_5^\text{MDA}$, column $\hat{M}^\text{DA}$ and column $\hat{M}^\text{MDA}$,  it is obvious that the attention-drift problem is alleviated by the mask supervision and the network can focus more on the inharmonious region. In this way, the redundant information could be abandoned with the help of spatial attention.  Specifically, one can observe that the refined encoder feature $a_5^\text{DA}$ filtered by DA is not always focusing on the inharmonious region, and may be distracted by other regions (as shown in row two, four, and seven), thereby leading to a wrong decision. By contrast, thanks to the supervision from ground-truth mask, the refined encoder feature filtered by MDA could accurately attend the context information to the inharmonious region.

\section{More Visualization Results of Mask Prediction}\label{sec:visualization}
\begin{figure*}[!ht]
\centering
\includegraphics[scale=0.6]{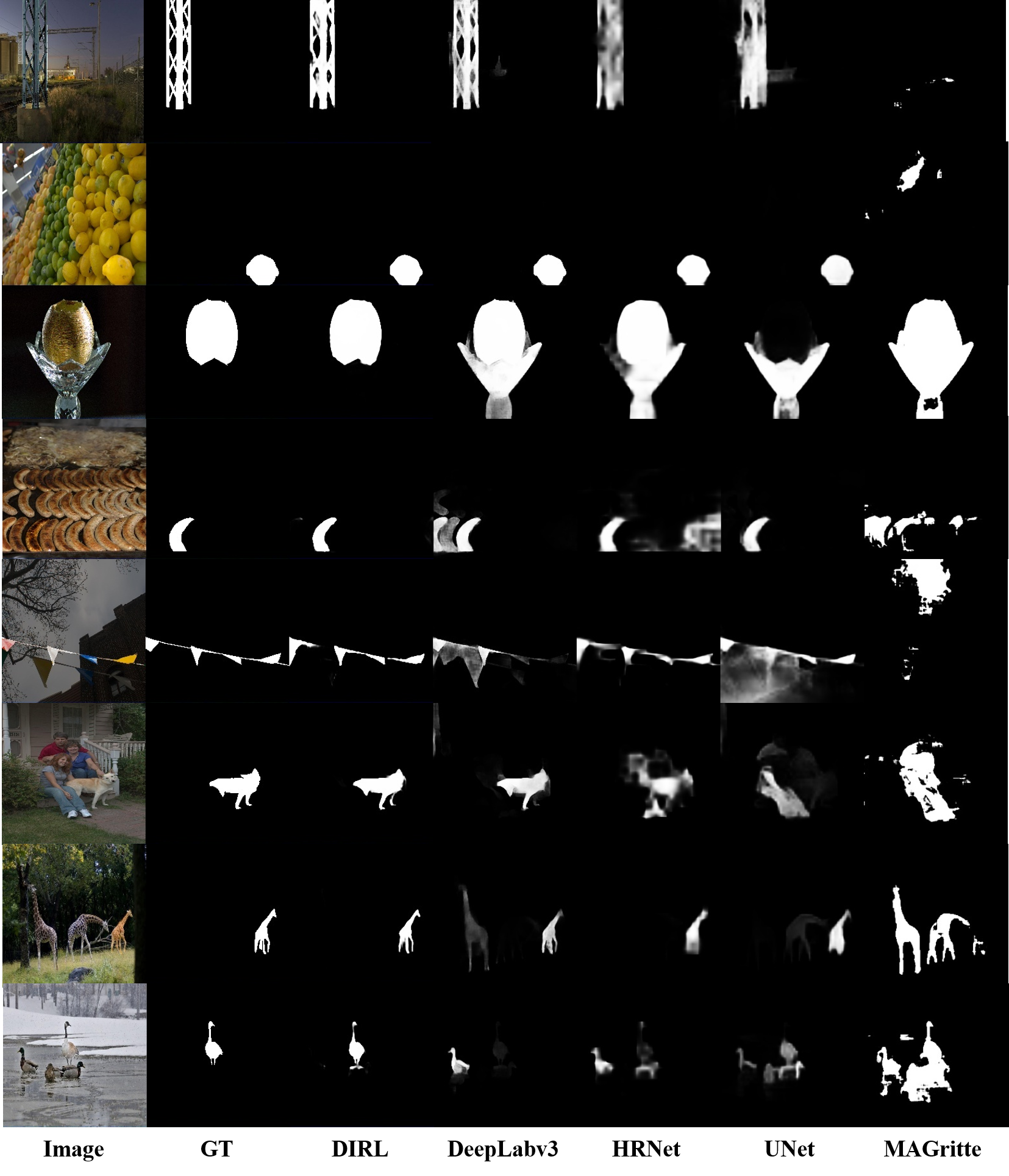}
\caption{More visualization results of different methods on the iHarmony4 dataset.}
\label{fig:mask2}
\end{figure*}

\begin{figure*}[!ht]
\centering
\includegraphics[scale=0.6]{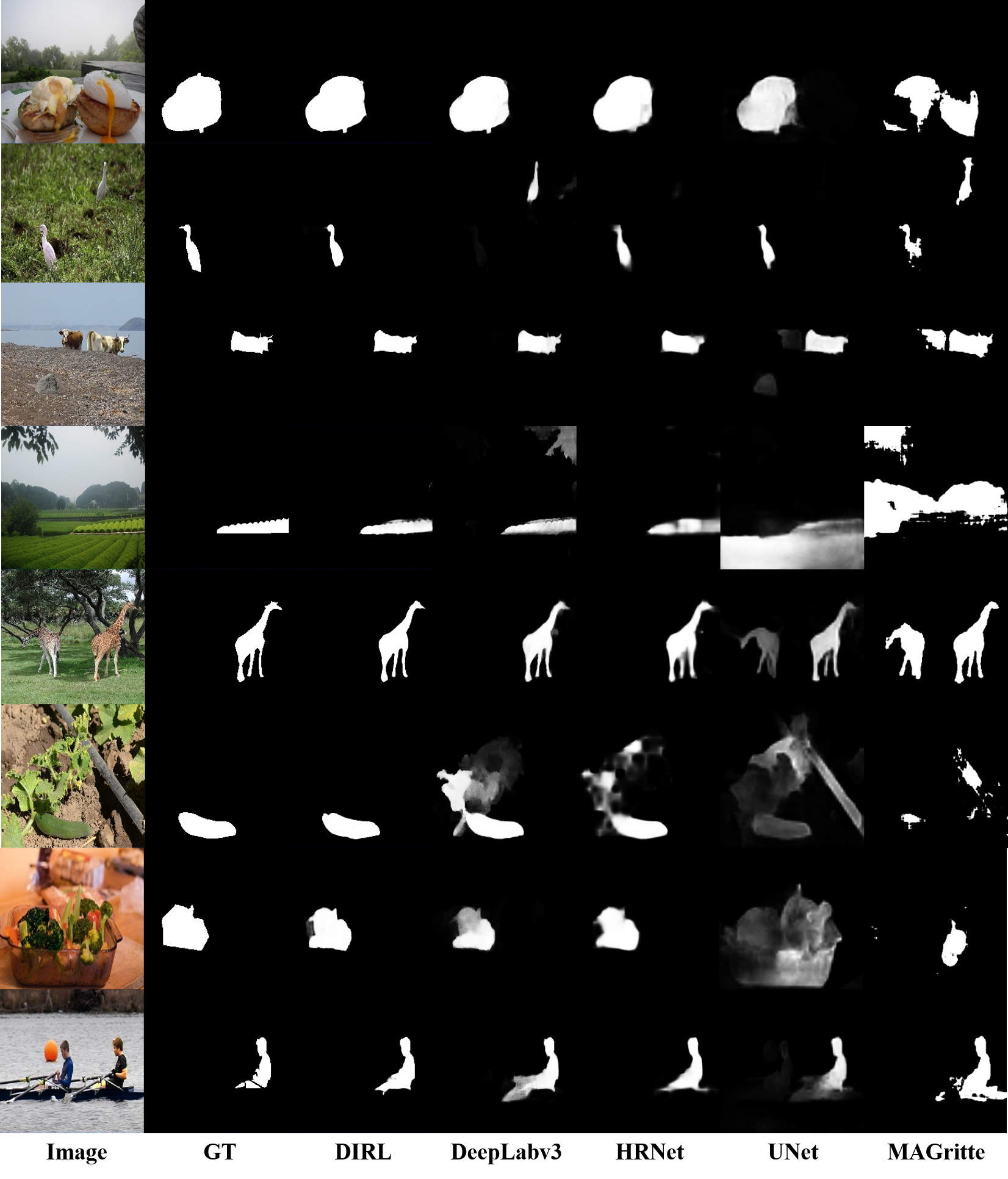}
\caption{More visualization results of different methods on the iHarmony4 dataset.}
\label{fig:mask3}
\end{figure*}

We show more examples of the inharmonious region masks predicted by different methods in Fig.~\ref{fig:mask2} and Fig.~\ref{fig:mask3}. In the second row in Fig.~\ref{fig:mask2}, most methods are able to discover the inharmonious region. In contrast, all of the baselines fail to capture the inharmonious region in the challenging background scene (see the last row in Fig.~\ref{fig:mask2}). In addition, our DIRL network is capable of clearly delineating the region boundary while other baselines result in blurry boundaries accompanied with artifacts (see Fig.~\ref{fig:mask2} and Fig.~\ref{fig:mask3}).
\bibliographystyle{IEEEbib}
\bibliography{supp}